\documentclass[sigconf]{acmart}
\usepackage{multirow,enumitem}
\usepackage{balance}

\AtBeginDocument{%
  \providecommand\BibTeX{{%
    \normalfont B\kern-0.5em{\scshape i\kern-0.25em b}\kern-0.8em\TeX}}}

\copyrightyear{2023}
\acmYear{2023}
\setcopyright{acmlicensed}
\acmConference[MM '23]{Proceedings of the 31st ACM International Conference on Multimedia}{October 29-November 3, 2023}{Ottawa, ON, Canada}
\acmBooktitle{Proceedings of the 31st ACM International Conference on Multimedia (MM '23), October 29-November 3, 2023, Ottawa, ON, Canada}
\acmPrice{15.00}
\acmDOI{10.1145/3581783.3611849}
\acmISBN{979-8-4007-0108-5/23/10}

\newcommand{\czq}[1]{\textcolor{black}{#1}}
\newcommand{\zj}[1]{\textcolor{black}{#1}}

\begin{document}
\title{Improving Anomaly Segmentation with Multi-Granularity Cross-Domain Alignment}
\author{Ji Zhang}
\affiliation{
  \institution{Southwest Jiaotong University, }
  \city{Chengdu}
  \country{China}
  }
\affiliation{
  \institution{Engineering Research Center of Sustainable Urban Intelligent Transportation, Ministry of Education}
  \city{}
  \country{China}
  }
\email{jizhang901@gmail.com}
\author{Xiao Wu}
\authornote{Corresponding authors: Xiao Wu and Zhi-Qi Cheng}
\authornotemark[1]
\affiliation{
  \institution{Southwest Jiaotong University, }
  \city{Chengdu}
  \country{China}
  }
\affiliation{
  \institution{Engineering Research Center of Sustainable Urban Intelligent Transportation, Ministry of Education}
  \city{}
  \country{China}
  }
\email{wuxiaohk@gmail.com}

\author{Zhi-Qi Cheng}
\authornotemark[1]
\affiliation{
    \institution{Language Technologies Institute, School of Computer Science, Carnegie Mellon University}
    \city{Pittsburgh}
    \country{United States}
}
\email{zhiqic@cs.cmu.edu}

\author{Qi He}
\affiliation{
  \institution{Southwest Jiaotong University, }
  \city{Chengdu}
  \country{China}
  }
\affiliation{
  \institution{Engineering Research Center of Sustainable Urban Intelligent Transportation, Ministry of Education}
  \city{}
  \country{China}
  }
\email{qihe96@gmail.com}
\author{Wei Li}
\affiliation{
  \institution{Southwest Jiaotong University, }
  \city{Chengdu}
  \country{China}
  }
\affiliation{
  \institution{Engineering Research Center of Sustainable Urban Intelligent Transportation, Ministry of Education}
  \city{}
  \country{China}
  }
\email{liwei@swjtu.edu.cn}
\renewcommand{\shortauthors}{Ji Zhang, Xiao Wu, Zhi-Qi Cheng, Qi He, \& Wei Li}

\begin{abstract}
Anomaly segmentation plays a pivotal role in identifying atypical objects in images, crucial for hazard detection in autonomous driving systems. While existing methods demonstrate noteworthy results on synthetic data, they often fail to consider the disparity between synthetic and real-world data domains. Addressing this gap, we introduce the Multi-Granularity Cross-Domain Alignment (MGCDA) framework, tailored to harmonize features across domains at both the scene and individual sample levels. Our contributions are twofold: 
\textit{i) We present the Multi-source Domain Adversarial Training module.} This integrates a multi-source adversarial loss coupled with dynamic label smoothing, facilitating the learning of domain-agnostic representations across multiple processing stages.
\textit{ii) We propose an innovative Cross-domain Anomaly-aware Contrastive Learning methodology.} This method adeptly selects challenging anchor points and images using an anomaly-centric strategy, ensuring precise alignment at the sample level.
Extensive evaluations of the Fishyscapes and RoadAnomaly datasets demonstrate MGCDA's superior performance and adaptability. Additionally, its ability to perform parameter-free inference and function with various network architectures highlights its distinctiveness in advancing the frontier of anomaly segmentation.
\end{abstract}

\ccsdesc[500]{Computing methodologies~Image segmentation}

\keywords{Anomaly segmentation; Domain adversarial training}
\maketitle
\begin{figure}[t!]
\includegraphics[width=0.9\linewidth]{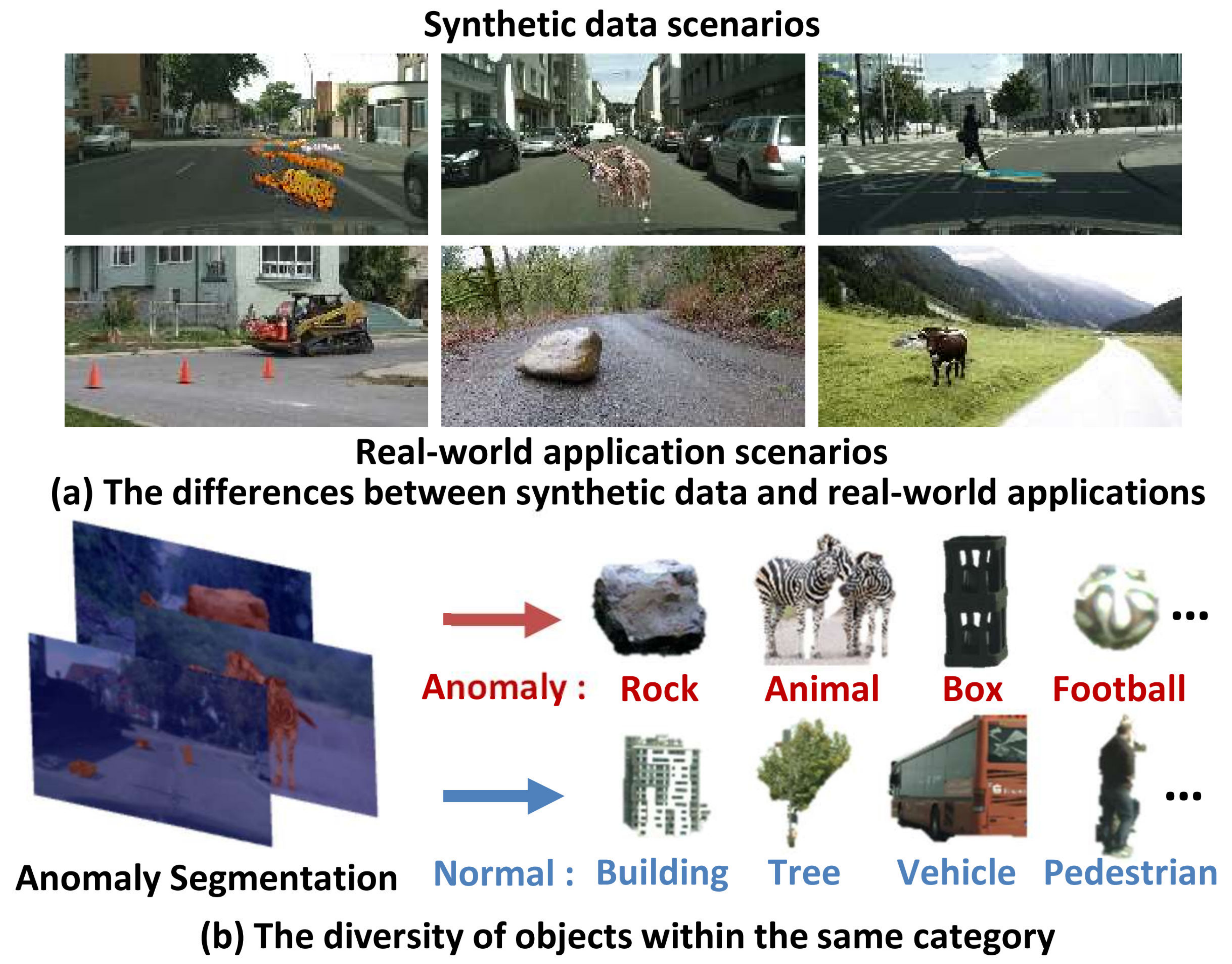}
\centering
\vspace{-0.1in}
\caption{\small The challenges primarily stem from the discrepancies between synthetic datasets and real-world environments, coupled with the inherent diversity of objects even within a single category. The anomaly segmentation datasets, such as LostAndFound \cite{ijcv_Fishyscapes} and RoadAnomaly \cite{ICCV_Imre}, span a more expansive range of settings, from suburban landscapes to expansive plains and untouched wilderness.}
\vspace{-0.15in}
\label{fig:problem}
\end{figure}

\section{Introduction}
\label{sec:intro}
The deep learning techniques have considerably propelled advancements in semantic segmentation for autonomous driving applications \cite{CVPR_FCN, PAMI_DeepLab, MM_RTSS, SWNet, mm_c, he2023damo,robust,li2023longshortnet,lan2023procontext}. \textit{However, one constant challenge arises from the diverse and versatile nature of anomalous objects, which are typically not part of the training datasets.} Such anomalies in open-world scenarios, like unexpected animals or obstructions, significantly compromise the safety of autonomous vehicles. To mitigate this, the concept of anomaly segmentation has been introduced \cite{ICCV_Imre}, acting as a valuable supplement to traditional semantic segmentation methods. By integrating anomaly segmentation, autonomous vehicles can adeptly detect and recognize unusual obstacles on the road, thereby enhancing their overall safety and dependability.

\begin{figure*}[t]
\includegraphics[width=0.9\linewidth]{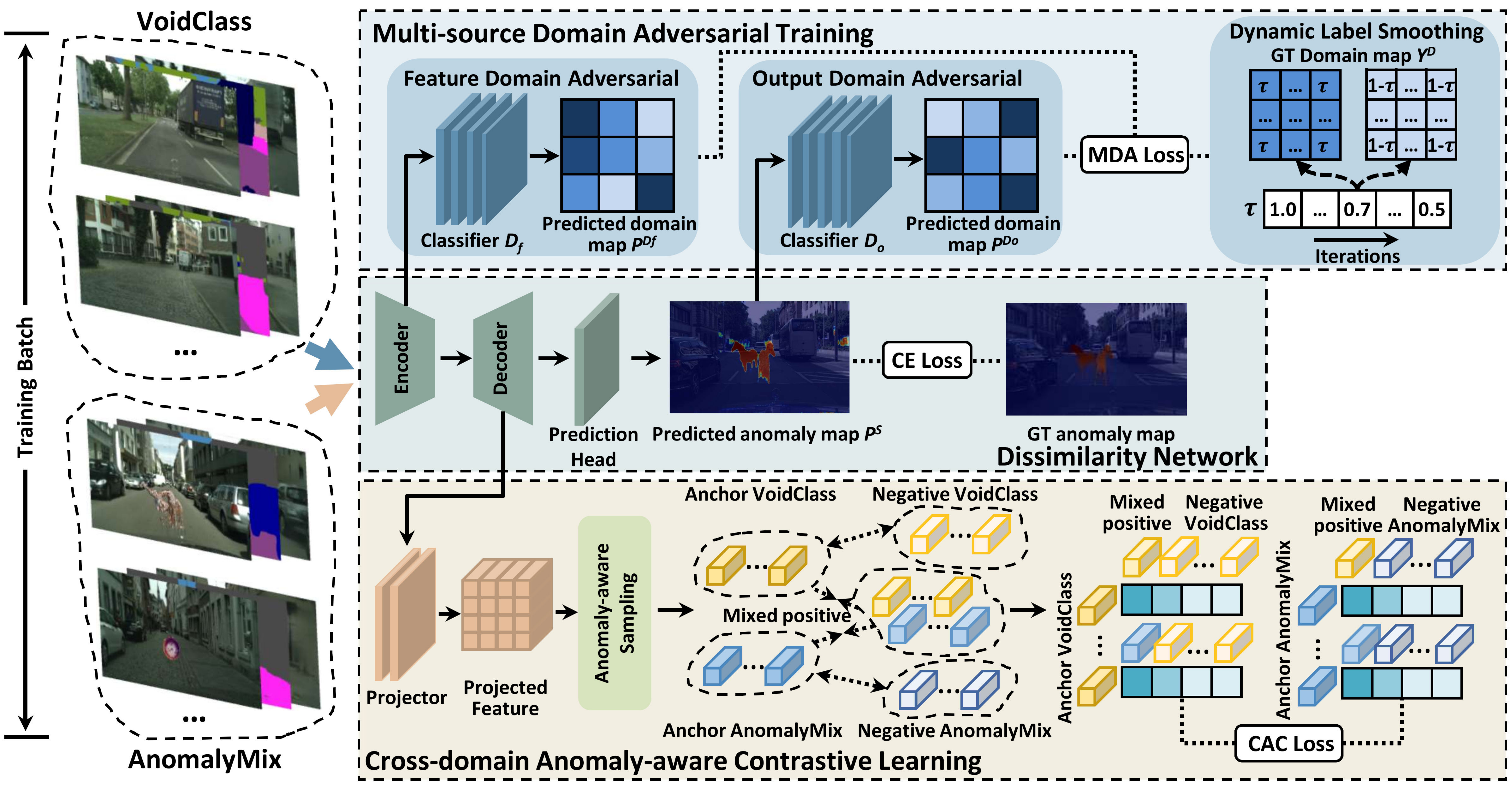}
\centering
\vspace{-0.1in}
\caption{\small The MGCDA framework is composed of two primary components: the Multi-source Domain Adversarial Training (MDAT) module and the Cross-domain Anomaly-aware Contrastive Learning (CACL) method. The MDAT module conducts domain adversarial training across multiple data sources through two stages within the discriminative network. Additionally, a dynamic label smoothing approach is integrated to adjust domain label values adaptively. The CACL method determines the contrastive loss based on pixels from cross-domain images and utilizes an anomaly-aware sampling approach to effectively select anchor, positive, and negative samples.}
\vspace{-0.1in}
\label{fig:ov}
\end{figure*}

\textit{Anomaly data are notably limited and challenging to gather, resulting in insufficient datasets for robust deep model training. }To circumvent this limitation, prior anomaly segmentation approaches \cite{NIPS_Bayesian, NIPS_ooDAA, ICCV_Imre, CVPR_syn, ICLR_OE, ECCV_pebal} have synthesized training data by leveraging existing semantic segmentation datasets, such as those found in \cite{CVPR_cityscape}. This synthetic data is typically generated by labeling rare objects as anomalies or by extracting anomalous objects from outlier datasets and superimposing them onto original images \cite{CVPR_syn,ECCV_pebal}.

\textit{However, a significant domain disparity and pronounced data variability exist between this synthetic training data and real-world scenarios.} As illustrated in Fig. \ref{fig:problem}, while training data predominantly captures urban street scenes, the anomaly segmentation datasets—like LostAndFound \cite{ijcv_Fishyscapes} \& RoadAnomaly \cite{ICCV_Imre}—encompass a broader spectrum of environments, from suburban areas to plains and wilderness regions. Consequently, the performance of methods trained on synthetic data is often suboptimal when confronted with real-world scenarios. This generalization gap is particularly concerning for autonomous driving applications, where models must operate seamlessly across diverse settings.

\textit{Moreover, another crucial issue is that anomalous objects are usually diverse and versatile, posing challenges for anomaly segmentation models to learn compact sample distributions.} In contrast to semantic segmentation, where objects within a category share resemblances, anomaly segmentation faces substantial intra-class differences. Positive samples in this context denote untrained anomalous objects, such as rocks, animals, boxes, and footballs. Conversely, negative samples encompass categories familiar to the model, like buildings, trees, vehicles, and pedestrians. This object diversity induces discrepancies in the distribution of sample representations between synthetic data and real-world contexts. \textit{Therefore, addressing both scene disparities and distribution variations becomes imperative when transitioning from synthetic data to real-world applications.}

\textit{To address these challenges, we introduce the Multi-Granularity Cross-Domain Alignment (MGCDA) framework, depicted in Fig. \ref{fig:ov}, designed to bolster the generalization capability of anomaly segmentation. }The framework combines the Multi-source Domain Adversarial Training (MDAT) module and the Cross-domain Anomaly-aware Contrastive Learning (CACL) method. Their collective aim is to reconcile disparities between synthetic data and real-world scenarios, considering both scene and sample distribution aspects. Initially, a dissimilarity network \cite{CVPR_syn} is employed to discern anomalous regions by contrasting original images with their reconstructions. Subsequently, MDAT is devised to promote the acquisition of domain-agnostic features. A distinct domain adversarial loss is introduced to bridge the feature distribution gap across training datasets. This loss operates in dual phases within the network to counteract gradient vanishing issues. Furthermore, a dynamic label smoothing approach is integrated to counteract adversarial instability during the convergence of multi-domain training data distributions. Lastly, to harmonize sample distribution disparities, the CACL method calculates pixel-wise contrastive losses, incorporating anchor and sample selection from cross-domain data.

\textit{Furthermore, the framework incorporates an anomaly-aware sampling mechanism, meticulously choosing challenging samples and anchors based on discriminator network predictions. }The amalgamation of the MDAT module and CACL method within the framework facilitates the training of a more robust model, culminating in enhanced precision in anomaly segmentation. A notable advantage of the proposed framework is its capacity to bolster the model's adaptability to unfamiliar scenes without augmenting the inference parameters. Additionally, the design of the framework is versatile, allowing it to seamlessly integrate with a wide array of anomaly segmentation techniques, serving as a modular addition. Overall, the key contributions of this work are:
\begin{itemize}
\item The inception of \textit{the Multi-Granularity Cross-Domain Alignment (MGCDA) framework}, aimed at bolstering the adaptability of anomaly segmentation methods to real scenarios. The framework's efficiency is highlighted by its parameter-free inference stage, and its design facilitates seamless integration with established anomaly segmentation architectures.

\item The introduction of \textit{multi-source domain adversarial training} to bridge scene-level variations. It incorporates a domain adversarial loss to nurture domain-neutral features, with a dynamic label smoothing approach adjusting to shifting feature distributions.

\item The formulation of \textit{a unique cross-domain anomaly-aware contrastive learning strategy}, focusing on aligning sample representations. \textit{An efficient anomaly-aware sampling technique is employed, optimizing the selection of anchors and samples based on anomaly predictions.}

\item Empirical validation on the Fishyscapes benchmark \cite{ijcv_Fishyscapes} and the RoadAnomaly \cite{ICCV_Imre} dataset underscores the framework's \textit{leading-edge performance against prevailing benchmarks}.
\end{itemize}

\section{Related Works}
\label{sec:related_work}
\subsection{Anomaly Segmentation}
Current anomaly segmentation techniques \cite{NIPS_Bayesian, NIPS_ooDAA, ICCV_Imre, CVPR_syn, ICLR_OE, ECCV_pebal} predominantly fall into three categories: \textit{1)~uncertainty, 2)~reconstruction, and 3)~outlier exposure} based approaches.

\noindent \textbf{Uncertainty-based Methods.}~These techniques \cite{NIPS_Bayesian, NIPS_ooDAA} discern anomalies by pinpointing regions with high predictive uncertainty. Initial approaches \cite{ICLR_deue,ICLR_cccue,ICCV_wvad} gauged uncertainty using softmax prediction values, proving effective for image-level tasks. However, they often misjudge pixels around anomalous object boundaries. To rectify this, boundary suppression techniques were pioneered by \cite{ICCV_sml}, and visual feature disparities were employed \cite{CVPRW_Drood} to delineate anomalies. Although subsequent techniques \cite{NIPS_Bayesian,arxiv_eb,NIPS_ssue} estimated uncertainty on a per-pixel basis to counteract boundary effects, their accuracy in anomaly segmentation was found wanting.

\noindent \textbf{Outlier Exposure Methods}~The outlier exposure (OE) strategy \cite{ICLR_OE} sought to enhance anomaly detection by leveraging an auxiliary outlier dataset. Yet, methods rooted in OE \cite{arxiv_dood,arxiv_rmtl} often necessitated retraining on this auxiliary set, potentially undermining the original segmentation model's efficacy. An integration approach was adopted by \cite{ECCV_pebal}, where the auxiliary dataset was amalgamated with the original training set.

\noindent \textbf{3)~Reconstruction-based Methods.}~These methods \cite{ICCV_Imre, CVPR_syn} reconstruct input visuals and juxtapose them with the originals to identify anomalies. While early iterations \cite{MICCAI_ruas} utilized autoencoders for image regeneration, the suboptimal quality of the output curtailed their effectiveness \cite{ICCV_Imre}. Contemporary techniques \cite{ICCV_Imre,arxiv_edss} attempt to reconstruct inputs from semantic maps produced by segmentation networks. However, their performance is contingent on the dissimilarity network's capacity to discern disparities between original and regenerated visuals \cite{CVPR_syn}. By integrating uncertainty maps, \cite{CVPR_syn} amalgamated both uncertainty-based and reconstruction-based methods into a unified framework.

Despite the noteworthy achievements of these anomaly segmentation strategies, \textit{the domain disparity between synthetic training sets and authentic testing sets remains inadequately addressed}. This gap underscores the impetus for this work, emphasizing multi-domain synthetic training to bolster model adaptability across multifaceted environments.

\subsection{Domain Adversarial Training}
Drawing inspiration from generative adversarial networks, adversarial training techniques \cite{ICML_DANN,AAAI_DAST,CVPR_adpseg,arxiv_fcns,CVPR_plda} have garnered attention across diverse areas such as image classification, object detection, and semantic segmentation. Central to their appeal is their capacity to foster domain-agnostic features, thereby bolstering the model's adaptability across varied domains.

The seminal work, FCAN \cite{arxiv_fcns}, marked the introduction of adversarial training to semantic segmentation, focusing on aligning feature-level information. Subsequent efforts, as seen in \cite{CVPR_adpseg}, shifted attention to aligning features in the output space, proving to be more proficient than intermediate feature alignments. Emphasizing pixel-level adversarial learning, \cite{CVPR_plda} sought to isolate image features, training the segmentation model on unadulterated content data. However, a common thread among these techniques is their objective to transfer models from labeled source domains to unlabeled target domains, leveraging data from the latter.

In contrast, \textit{our endeavor seeks to hone the generalization prowess of models on authentic test datasets}, employing a plethora of labeled synthetic training data. This multiplicity of synthetic training data, bearing scene similarities, presents intricate challenges for the adversarial training paradigm.

\vspace{-3mm}
\subsection{Contrastive Learning}
Contrastive learning operates on the principle of narrowing distances between features of identical classes while widening those of distinct classes. Recently, unsupervised contrastive learning techniques \cite{ICML_simcl,CVPR_Moco,CVPR_ucl} have demonstrated notable advancements, outperforming earlier methods grounded in pretext tasks \cite{ECCV_pt1,ECCV_pt2}. 

For instance, a foundational approach \cite{ICML_simcl} employs contrastive learning, wherein augmented renditions of an original image serve as positive exemplars, while alternate images act as negatives. The innovative memory bank mechanism, as championed by \cite{CVPR_Moco}, provides an expanded reservoir of negative samples. In the realm of image classification, a supervised contrastive loss has been formulated \cite{arxiv_scl,hu2021subspace}.

Building on this, several techniques \cite{ECCV_mscscl,ICCV_SSCL,ICCV_less,ICCV_racs,cheng2019learning,cheng2022rethinking} have extended supervised contrastive loss to pixel-level semantic segmentation. For instance, \cite{ICCV_less} pre-trains the segmentation network using both intra-image and cross-image contrastive losses. Works like \cite{ICCV_SSCL,ICCV_racs} employ a memory bank to house pixel or region embeddings, aggregated by class. Multi-scale and cross-scale contrastive learning is the focus of \cite{ECCV_mscscl}. To bolster the domain adaptability of semantic segmentation models across datasets, \cite{ECCV_BCL} dynamically crafts pseudo labels for target domain data, aligning pixel-level features and class prototypes between target and source visuals.

In the specific realm of anomaly segmentation, both positive and negative samples manifest pronounced intra-class variances. \textit{This distinction becomes even more pronounced in cross-domain datasets}, presenting a significant challenge to existing pixel-level contrastive learning methodologies.

\section{Proposed MGCDA Framework}
\label{sec:method}
\subsection{Framework Overview}
\label{sec:framework}
Addressing the prevalent domain disparities between synthetic datasets and real-world contexts, we introduce the Multi-Granularity Cross-Domain Alignment (MGCDA) framework, visually represented in Fig. \ref{fig:ov}. We incorporate the dissimilarity network, as delineated in \cite{CVPR_syn}, to identify anomalous regions using input datasets, comprising original images, semantic mappings, and reconstructed visuals. The subsequent cross-entropy loss, derived from the predicted anomaly maps and their ground truth counterparts, steers the network's optimization.

Our Multi-source Domain Adversarial Training (MDAT) technique aims to reconcile scene-level domain variances. \textit{Central to MDAT is a domain adversarial loss tailored for multi-source alignment and a dynamic label smoothing mechanism to bolster training consistency.} Recognizing the potential for gradient decay in more intricate network structures, we strategically deploy MDAT across both encoder and decoder phases. Additionally, the Cross-domain Anomaly-aware Contrastive Learning (CACL) approach is instituted to synchronize cross-domain sample representations, ensuring a tight intra-class and distinct inter-class distribution. For computational efficiency, an Anomaly-aware Sampling mechanism is embedded within the CACL, optimizing the selection of challenging instances and anchors.

\textit{Detailing the structure, the dissimilarity network comprises three key segments: an encoder, a decoder, and a prediction head.} As illustrated in Fig.~\ref{fig:ov}, the VGG network, pre-trained as per \cite{ICLR_vgg}, encodes both original and regenerated visuals, while a rudimentary CNN caters to the semantic map encoding. For each tier of the feature pyramid, the input, synthesis, and semantic feature maps undergo fusion via a \(1 \times 1\) convolution. The fused feature map is then juxtaposed with the uncertainty map to highlight regions of elevated uncertainty. Layer-wise, the decoder interprets each feature map, integrating it with the corresponding superior-level map to produce the anomaly segmentation prediction. To safeguard semantic integrity, spatial-aware normalization \cite{CVPR_sis} is also applied.

\subsection{Training Data Synthesis}
For the purposes of this study, we employ the data generation techniques detailed in \cite{CVPR_syn,ECCV_pebal}, focusing on the synthesis of training data for VoidClass and AnomalyMix. The VoidClass dataset is constructed by designating objects within the 'void' category of ground-truth semantic maps, such as those in Cityscapes \cite{CVPR_cityscape}, as anomalies. While the anomalies and scenes within VoidClass maintain stylistic congruence, the range and diversity of anomalous objects are inherently constrained by the constituents of the void category.

Conversely, the AnomalyMix dataset is synthesized by curating objects from the COCO dataset \cite{ECCV_Coco} as anomalies and subsequently superimposing them onto images drawn from the Cityscapes training set. This approach infuses AnomalyMix with a diverse array of anomalous objects, varying in scale and type. However, a stylistic disparity often emerges between the anomalous entities and the backdrop scenes. Recognizing their individual limitations, it becomes evident that VoidClass and AnomalyMix datasets, with their distinct domain variances, can be mutually complementary. In alignment with \cite{CVPR_syn}, ground-truth anomaly maps are crafted by marking the anomalous object regions as 1, the void category (excluding regions designated as anomalies) as 255, and all remaining areas as 0. For clarity, the datasets from VoidClass and AnomalyMix are denoted as \(I_V\) and \(I_A\), respectively.

\subsection{Multi-Source Domain Adversarial Training}
\label{sec:mmdat}
While numerous domain adversarial training techniques \cite{AAAI_DAST,arxiv_fcns,CVPR_adpseg,CVPR_plda} in dense prediction tasks utilize unlabeled target domain data to bridge the domain gap from source to target, their primary focus has been solely on the target domain data for adversarial loss computation. However, in the unique context of anomaly segmentation, the training sets can be synthesized through diverse methodologies, leaving real-world test sets often uncharted.

Particularly, addressing this, our Multi-source Domain Adversarial Training (MDAT) approach harnesses multiple source domain datasets, aiming for enhanced adaptability to unfamiliar target domains. The primary thrust of MDAT's multi-source domain adversarial loss is to harmonize features across varied source domains, fostering more congruent distributions instead of merely aligning the target to source domain features.

As training progresses, the convergence of feature distributions from disparate source domains can challenge traditional rigid domain label assignment strategies, possibly destabilizing model training \cite{ARXIV_els}. Our solution is a dynamic label smoothing strategy, designed to bolster training stability through adaptive domain label assignments. Moreover, recognizing potential gradient decay challenges in deeper network structures, we've instituted a multi-stage domain adversarial strategy, applied distinctly within both the encoder and decoder network phases.

In the MDAT schema, features \( F \) from the encoder and the anomaly map \( P^{S} \) from the dissimilarity network are channeled into the feature domain classifier \( {D_f} \) and the output domain classifier \( {D_o} \), producing the predicted domain maps \( P^{D_f} \) and \( P^{D_o} \). Concurrently, ground-truth domain maps \( Y^{D} \) are crafted by assigning domain label values of \( \tau \) to VoidClass and \( 1 - \tau \) to AnomalyMix. The value of \( \tau \) dynamically adapts during training, as described by:
\begin{equation}
\tau = \tau_{b} - (\tau_{b} - 0.5)\frac{i}{max\_i},
\end{equation}
where \( \tau_{b} \) represents an initial hyperparameter, and \( i \) denotes the current training iteration count. 

Given the predicted domain map \( P^{D_f} \) and the ground-truth domain map \( Y^{D} \), the multi-source domain adversarial loss for the feature domain classifier is formulated as:
\begin{equation}
\mathcal L_{MDA}^{D_f}=\frac{1}{N}\sum_{i=1}^N(P^{D_f}_V - Y^{D}_A)^2 + (P^{D_f}_A - Y^{D}_V)^2,
\end{equation}
where \( N \) signifies the number of examples, while \( P^{D_f}_V \) and \( P^{D_f}_A \) are domain maps predicted from images in VoidClass and AnomalyMix, respectively. The multi-source domain adversarial loss for the output domain classifier, \( \mathcal L_{MDA}^{D_o} \), follows a similar computation.

Elaborating on the architecture, the feature domain classifier comprises a sequence of four convolution layers, represented as \( \{C_{ic/2, 3}^1-C_{ic/4, 3}^1-C_{ic/8, 3}^1-C_{1, 3}\} \). Here, \( C_{oc, k}^s \) signifies a convolution layer with a kernel of \( k \) and stride of \( s \), with \( oc \) and \( ic \) denoting the output and input channels, respectively. Post every convolution layer, except the terminal one, a ReLU operation is applied. In a parallel vein, the output domain classifier adheres to the structure \( \{C_{32, 3}^1-C_{64, 3}^1-C_{128, 3}^1-C_{256, 3}^1-C_{1, 3}\} \).

\subsection{Cross-domain Anomaly-aware Contrastive Learning}
\label{sec:cacl}
The inherent complexity of anomaly segmentation tasks is compounded by the sheer diversity of anomalous objects. This diversity often disrupts the formation of a structured sample distribution, pivotal for effective modeling. The chasm between synthetic data representations and real-world application scenarios further amplifies the challenge of aligning features of samples within similar categories.

In light of the successes demonstrated by contrastive learning in multiple tasks \cite{CVPR_Moco,ICML_simcl}, this work introduces the Cross-domain Anomaly-aware Contrastive Learning (CACL) approach. CACL seeks to tighten the distribution of features within identical categories while widening the gap between different categories. Within the CACL framework, each image pixel is treated as an individual sample. Pixels originating from anomalous regions are designated as positive samples, whereas those from standard regions are labeled as negative samples. However, the stark differences between positive and negative samples, particularly across various domains, could potentially dilute the effectiveness of the contrastive loss, as it may become inundated by a plethora of straightforward cases. To circumvent this, anchors are paired with positive samples from different domains and negative samples from the same domain, fostering robust contrastive learning on cross-domain data.

Given the granularity inherent in anomaly detection tasks, executing pixel-wise contrastive loss becomes computationally daunting. While several studies \cite{ICCV_racs,ICCV_less} have endeavored to mitigate this through sampling strategies or category prototype techniques, they tend to falter in anomaly segmentation due to pronounced intra-class differences. Addressing this gap, we devise an anomaly-aware sampling strategy, which judiciously selects samples and anchors based on anomaly segmentation outcomes, optimizing cross-domain pixel-wise contrastive loss computation. Drawing inspiration from the hard sample mining method \cite{CVPR_oh}, this strategy discerns between correctly predicted pixels (easy samples) and misclassified pixels (hard samples). To bolster successful contrastive learning and avert potential pitfalls into local optima, anchors are sourced from both hard and easy anomaly samples. Similarly, negative samples are meticulously curated from both hard and easy standard samples.

\begin{table*}[htbp]
\caption{\small Comparison with SOTA methods on Fishyscapes Leaderboard. The best and second \czq{are} shown in \textbf{bold} and \underline{underlined}, respectively.}
\vspace{-0.15in}
\begin{center}
\begin{tabular}{l|c|c|c|cc|cc}
\hline 
\multirow{2}{*}{Method} & \multirow{2}{*}{Venue} & \multirow{2}{*}{Re-training} & \multirow{2}{*}{OoD Data} & \multicolumn{2}{c|}{FS LostAndFound} & \multicolumn{2}{c}{FS Static}\cr
& & & & AP(\%)$\uparrow$ & FPR95(\%)$\downarrow$ & AP(\%)$\uparrow$ & FPR95(\%)$\downarrow$ \cr\hline \hline
MSP \cite{ICML_msp} & ICML’21 & $\times$ & $\times$ & 1.77 & 44.85 & 12.88 & 39.83 \cr
Entropy \cite{ICLR_deue} & ICLR’17 & $\times$ & $\times$ & 2.93 & 44.83 & 15.41 & 39.75 \cr
Density-Minimum \cite{ijcv_Fishyscapes} & IJCV’21 & $\times$ & $\times$ & 4.25 & 47.15 & 62.14 & 17.43 \cr
Image Resynthesis++ \cite{ICCV_Imre} & ICCV’19 & $\times$ & $\times$ & 5.70 & 48.05 & 29.60 & 27.13 \cr
SML \cite{ICCV_sml} & ICCV’21 & $\times$ & $\times$ & 31.05 & 21.52 & 53.11 & 19.64 \cr
\hline
Bayesian Deeplab \cite{arxiv_eb} & Arxiv’18 & $\checkmark$ & $\times$ & 9.81 & 38.46 & 48.70 & 15.05 \cr
\hline
Logistic Regression \cite{ijcv_Fishyscapes} & IJCV’21 & $\times$ & $\checkmark$ & 4.65 & 24.36 & 57.16 & 13.39 \cr
Synboost \cite{CVPR_syn} & CVPR’21 & $\times$ & $\checkmark$ & 43.22 & 15.79 & 72.59 & 18.75 \cr
PEBAL \cite{ECCV_pebal} & ECCV’22 & $\times$ & $\checkmark$ & \underline{44.17} & 7.58 & \underline{92.38} & \underline{1.73} \cr
\hline
OoD training \cite{ijcv_Fishyscapes} & IJCV'21 & $\checkmark$ & $\checkmark$ & 10.29 & 22.11 & 45.00 & 19.40 \cr
Outlier Head \cite{GCPR_outlier} & GCPR'19 & $\checkmark$ & $\checkmark$ & 31.31 & 19.02 & \textbf{96.76} & \textbf{0.29} \cr
Dirichlet DeepLab \cite{NIPS_dl} & NIPS'18 & $\checkmark$ & $\checkmark$ & 34.28 & 47.43 & 31.30 & 84.60 \cr
DenseHybrid \cite{ECCV_dh} & ECCV’22 & $\checkmark$ & $\checkmark$ & 43.90  & \textbf{6.18} & 72.27 & 5.51 \cr
\hline
MGCDA & - & $\times$ & $\checkmark$ & \textbf{60.96} & \underline{6.66} & 74.97 & 17.71 \cr
\hline
\end{tabular}
\end{center}
\vspace{-0.1in}
\label{tab:fs}
\end{table*}
\begin{table}[htbp]
\caption{\small Comparison with SOTA methods on RoadAnomaly Dataset.}
\vspace{-0.15in}
\begin{center}
\begin{tabular}{l|c|cc}
\hline 
Method & Venue & AP(\%)$\uparrow$ & FPR95(\%)$\downarrow$ \cr
\hline \hline
Entropy \cite{ICLR_deue} & ICLR’17 & 16.97  & 71.10 \cr
SML\cite{ICCV_sml} & ICCV’21 & 17.52 & 70.70 \cr
MSP \cite{ICML_msp} & ICML’21 & 15.72 & 71.38 \cr
Max Logit \cite{ICML_msp} & ICML’21 & 18.98 & 70.48 \cr
Energy \cite{NIPS_energy} & NIPS'20 & 19.54 & 70.17 \cr
GMMSeg-DeepLabV3+ \cite{NIPS_gmmseg} & NIPS'22 & 34.42 & 47.90 \cr
Synboost \cite{CVPR_syn} & CVPR’21 & 38.21 & 64.75 \cr
PEBAL \cite{ECCV_pebal} & ECCV’22 & \underline{45.10} & \underline{44.58} \cr
\hline
MGCDA & - & \textbf{50.35} & \textbf{42.19} \cr
\hline
\end{tabular}
\end{center}
\vspace{-0.1in}
\label{tab:road}
\end{table}

Within CACL, features derived from the decoder transform a higher-dimensional space using a projector, enhancing their representational prowess. For each image \(I_i\) (\(i \in I^V \cup I^A\)) in the training batch, the anomaly-aware sampling strategy delineates anchor, positive, and negative sets. These projected features are first demarcated into the anomaly and standard regions in pixel space, leveraging the ground-truth anomaly map. Employing the predicted anomaly map, both regions are further subdivided into hard and easy segments. From these regions, samples are extracted to form anchor, positive, and negative sets. Through a meticulously orchestrated process, the cross-domain anomaly-aware contrastive loss is articulated as:
\begin{equation}
\begin{aligned}
\mathcal L_{CAC}= &-\frac{1}{|S^a_{V}|}\sum_{a \in S^a_{V}}\frac{1}{|S^p_{V \cup A}|}\sum_{p \in S^p_{V \cup A}}\frac{exp^{a \cdot p / \alpha}}{exp^{a \cdot p / \alpha} + \sum_{n \in S^n_V}exp^{a \cdot n / \alpha}}
\\&-\frac{1}{|S^a_{A}|}\sum_{a \in S^a_{A}}\frac{1}{|S^p_{V \cup A}|}\sum_{p \in S^p_{V \cup A}}\frac{exp^{a \cdot p / \alpha}}{exp^{a \cdot p / \alpha} + \sum_{n \in S^n_A}exp^{a \cdot n / \alpha}}
\vspace{-0.2cm}
\label{nce_loss}
\end{aligned}
\end{equation}
where \( \alpha \) represents a temperature parameter. The projector is realized through two \( 1 \times 1 \) convolutions with ReLU, delineated as \( \{C_{64, 1}^1-C_{128, 1}^1\} \).

\subsection{Loss Function}
The proposed framework harnesses a combination of various loss components for effective optimization. Specifically, the cross-entropy loss \(\mathcal L_{CE}\), the multi-source domain adversarial losses \(\mathcal L_{MDA}^{D_f}\) and \(\mathcal L_{MDA}^{D_o}\), along with the cross-domain anomaly-aware contrastive loss \(\mathcal L_{CAC}\), collectively drive the training process. The aggregate loss function can be succinctly represented as:
\begin{equation}
\mathcal L_{Total} = \mathcal L_{CE} + \lambda_f \mathcal L_{MDA}^{D_f} + \lambda_o \mathcal L_{MDA}^{D_o} + \lambda_c \mathcal L_{CAC},
\end{equation}
where the terms \(\lambda_f\), \(\lambda_o\), and \(\lambda_c\) serve as weight coefficients that modulate the relative contributions of the respective loss components to the final loss.

\section{Experiments}
\label{sec:experiments}
\subsection{Datasets and Evaluation Metrics}
We evaluate the proposed method on two prominent anomaly segmentation datasets: the RoadAnomaly dataset and the Fishyscapes benchmark, each offering unique scene compositions.

\noindent \textbf{Fishyscapes} \cite{ijcv_Fishyscapes} is a benchmark specifically designed to assess anomaly detection within urban driving scenarios. It splits into two subsets as follows:
\begin{itemize}
    \item \textit{FS LostAndFound}: Contains genuine roadway anomalies with a validation set of 100 images and a test set comprising 275 images.
    \item \textit{FS Static}: Integrates anomalous objects from Pascal VOC into Cityscapes validation images, resulting in a validation set of 60 images and a test set of 1000 images.
\end{itemize}

\noindent \textbf{RoadAnomaly} \cite{ICCV_Imre} is a collection of 60 real-world images sourced online, depicting unexpected objects near vehicles, like wildlife, debris, and construction machinery. Detailed pixel-level annotations identify the anomalies. Distinguished from Fishyscapes, RoadAnomaly encompasses a broader spectrum of anomaly sizes and forms. All images are standardized to a resolution of \(1280 \times 720\).

\noindent \textbf{Evaluation Metrics}.~In alignment with established practices, we adopt the evaluation metrics from the Fishyscapes benchmark \cite{ijcv_Fishyscapes}:
\begin{itemize}
    \item \textit{Average Precision (AP)}: Measures the model's pixel-wise classification prowess, indicating its precision.
    \item \textit{FPR95}: Denotes the false positive rate at a 95\% true positive rate, assessing the model's reliability for safety-critical applications.
\end{itemize}

\begin{figure*}[t]
\includegraphics[width=\linewidth]{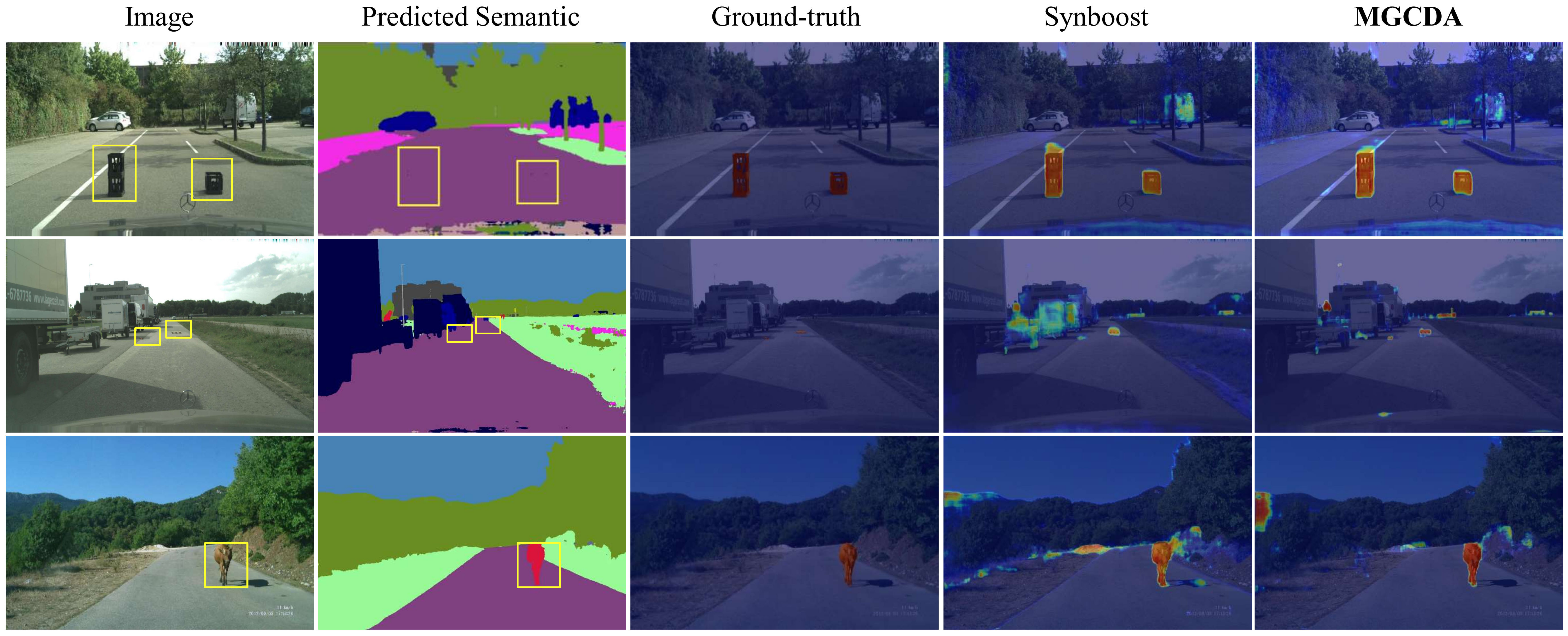}
\centering
    \vspace{-0.15in}
\caption{\small \zj{Visualization of the predicted anomaly maps on Fishyscapes benchmark and RoadAnomaly dataset. Anomaly objects in images are highlighted with a yellow box.}}
\vspace{-0.1in}
\label{fig:viss}
\end{figure*}

\subsection{Implementation Details}
To ensure an equitable evaluation of our proposed framework, we adhere to the network configurations outlined in \cite{CVPR_syn}, which was trained on the Cityscapes dataset \cite{CVPR_cityscape}. Here are the specific settings and configurations employed:
\begin{itemize}
    \item \textit{Training Duration}: The dissimilarity network undergoes training for 50 epochs.
    \item \textit{Optimizer}: We utilize the Adam optimizer \cite{ICLR_adam} with an initial learning rate set at 0.0001.
    \item \textit{Learning Rate Schedule}: A polynomial schedule modulates the learning rate, employing a power factor of 0.99.
    \item \textit{Batch Configuration}: The batch size is configured at 8, comprising 4 samples from each domain.
    \item \textit{Data Augmentation}: Training data undergo augmentation through horizontal flipping and normalization, using the mean and standard deviation values from the Imagenet dataset \cite{IJCV_imagenet}.
    \item \textit{Domain Labeling}: The initial domain label value, denoted as \(\tau_{base}\), is set at 1.
    \item \textit{Loss Weights}: The respective weights for the multi-source domain adversarial losses are: \(\lambda_f = 0.04\), \(\lambda_o = 0.06\), and \(\lambda_c = 0.1\).
    \item \textit{Sampling Parameters}: Through hyper-parameter tuning, the optimal anchor count \(N^{a}\) is determined to be 50. Based on empirical observations, the positive sample count \(N^{p}\), hard negative sample count \(N^{n}_h\), and easy negative sample count \(N^{n}_e\) are respectively set to twice, twice, and six times the anchor count \(N^{a}\).
\end{itemize}

\subsection{Comparisons with State-of-the-Art Methods}
We evaluate our proposed method against leading state-of-the-art approaches on three datasets: FS LostAndFound, RoadAnomaly, and FS Static. In line with the standards set by \cite{ijcv_Fishyscapes}, we categorize these methods in Table \ref{tab:fs} based on their re-training requirements and their use of additional Out-of-Distribution (OoD) data.

In the FS LostAndFound dataset, as detailed in Table \ref{tab:fs}, our framework demonstrates a substantial enhancement, registering a 16.79\% uptick in AP performance coupled with a low FPR95 of 6.66\%. Prior methodologies \cite{CVPR_syn,ECCV_pebal} predominantly trained their models on synthetic datasets, which, owing to domain discrepancies with FS LostAndFound, often struggled to achieve top-tier results. In contrast, our approach fortifies the generalization prowess of the baseline model \cite{CVPR_syn} through multi-source domain adversarial training and cross-domain anomaly-aware contrastive learning, translating to superior outcomes on FS LostAndFound.

For the RoadAnomaly dataset, detailed in Table \ref{tab:road}, our approach shines brightly. Given that this dataset, sourced from the internet, offers a more eclectic mix of scenarios and objects than FS LostAndFound, it stands as a testament to the robustness of our framework. The results indicate a 12.14\% improvement in AP and a 22.56\% reduction in FPR95 over the baseline, marking the best performance in both metrics. This underscores the universal applicability of our framework across varied real-world test sets.

For the FS Static dataset, which is synthesized by overlaying out-of-distribution objects onto the Cityscapes dataset \cite{CVPR_cityscape}, it shares significant similarities with synthetic training data. The apex performance on FS Static is claimed by \cite{GCPR_outlier}, which undertakes a comprehensive retraining of their semantic segmentation network using synthetic data. Another noteworthy contender, \cite{ECCV_pebal}, refines the concluding classification block with synthetic data, achieving commendable results. Impressively, without any retraining, our framework still secures competitive outcomes. This suggests that fostering domain-neutral features and cultivating compact sample distributions can bolster a model's efficacy even on identical domain data.

In summation, while we may not clinch the top spot in every metric, our framework's accomplishments are undeniable. It surpasses other techniques in achieving three of the best results and one runner-up performance on real-world datasets. As visualized in Fig. \ref{fig:viss}, our model's anomaly predictions are juxtaposed with those of Synboost \cite{CVPR_syn}. Both models exhibit anomaly maps with pronounced scores for anomalous pixels. However, our model trumps with fewer false positives and negatives, aptly detecting even subtle anomalies. The fact that our model also excels on synthetic data underscores its enhanced generalization capabilities.
\begin{table*}[htbp]
\caption{\small Ablation studies on Fishyscapes benchmark validation set and RoadAnomaly dataset.}
\vspace{-0.15in}
\begin{center}
\begin{tabular}{c|ccc|ccc|cc|cc|cc}
\hline 
Training & \multicolumn{3}{c|}{MDAT} & \multicolumn{3}{c|}{CACL}& \multicolumn{2}{c|}{FS LostAndFound} & \multicolumn{2}{c|}{FS Static} & \multicolumn{2}{c}{RoadAnomaly}\cr
Data & ODA & FDA & DLS & PCL & CPCL & AS & AP(\%)$\uparrow$ & FPR95(\%)$\downarrow$ & AP(\%)$\uparrow$ & FPR95(\%)$\downarrow$ & AP(\%)$\uparrow$ & FPR95(\%)$\downarrow$ \cr\hline \hline
VoidClass & & & & & & & 60.58 & 31.02 & 66.44 & 25.59 & 38.21 & 64.75 \cr
AnomalyMix & & & & & & & 18.45 & 40.52 & 77.87 & 19.41 & 33.82 & 65.45 \cr
Mix & & & & & & & 56.71 & 37.90 & 67.61 & 22.17 & 34.45 & 65.81 \cr
\hline
Multi-domain & $\checkmark$ & & & & & & 67.29 & 38.20 & 74.63 & 23.36 & 43.01 & 55.59 \cr
Multi-domain & $\checkmark$ & $\checkmark$ & & & & & 68.99 & 25.39 & 76.56 & 21.95 & 45.24 & 56.33 \cr
Multi-domain & $\checkmark$ & $\checkmark$ & $\checkmark$ & & & & 70.20 & 25.76 & 78.05 & 26.12 & 46.17 & 53.89 \cr
\hline
Multi-domain & $\checkmark$ & $\checkmark$ & $\checkmark$ & $\checkmark$ & & & 71.55 & 24.73 & 77.95 & 24.15 & 46.87 & 51.23 \cr
Multi-domain & $\checkmark$ & $\checkmark$ & $\checkmark$ & & $\checkmark$ & & 72.30 & 22.15 & 78.79 & \textbf{17.15} & 47.95 & 48.29 \cr
Multi-domain & $\checkmark$ & $\checkmark$ & $\checkmark$ & & $\checkmark$ & $\checkmark$ & \textbf{74.71} & \textbf{19.08} & \textbf{80.90} & 22.72 & \textbf{50.35} & \textbf{42.19} \cr\hline
\end{tabular}
\end{center}
\label{tab:ab}
\end{table*}

\subsection{Ablation Studies}
We dissect the impact of various components of our method using ablation studies. The experiments are performed on the Fishyscapes validation sets and the RoadAnomaly dataset, as discussed in Sec. \ref{sec:method}. The baseline method from \cite{CVPR_syn} serves as our reference point.

\subsubsection{\textbf{Influence of Extra Training Data}}
To delineate the impact of additional training data, the baseline model is trained using three different datasets. When trained solely on $VoidClass$ or $AnomalyMix$, the model's performance varies. Notably, while the baseline achieves commendable results on FS Static when trained on $AnomalyMix$, it underperforms on the FS LostAndFound and RoadAnomaly datasets. This can be attributed to the synthetic nature of the $AnomalyMix$ data, which creates a stark contrast between the styles of anomalies and their surrounding scenes. Consequently, the model becomes predisposed to identifying style discrepancies, adversely affecting its anomaly segmentation capabilities.

Furthermore, when data from both domains are amalgamated (denoted as $Mix$), the model's performance takes a hit compared to when trained exclusively on $VoidClass$. This highlights the underlying domain differences between datasets synthesized differently. Merely pooling the data might skew the model towards a particular domain, undermining its generalizability.

\subsubsection{\textbf{Discussion on the Multi-source Domain Adversarial Training Method}}

We undertake an in-depth analysis of the efficacy of the proposed multi-source domain adversarial training method by comparing three distinct network configurations:

\begin{itemize}
    \item (a) $Baseline+ODA$: Integrates the output domain adversarial module into the baseline.
    \item (b) $Baseline+ODA+FDA$: Adds the feature domain adversarial module to create a multi-stage adversarial training architecture.
    \item (c) $Baseline+MDAT$: Deploys the dynamic label smoothing strategy for on-the-fly domain label adjustments.
\end{itemize}

As evident from Table \ref{tab:ab}, the $Baseline+ODA$ configuration boasts an enhancement in AP performance on FS LostAndFound and RoadAnomaly datasets. However, there's a slight dip in the FS Static dataset. These results corroborate our hypothesis about the presence of domain disparities between synthetic and real-world data. By emphasizing generalizability, the model's performance in real-world scenarios is bolstered.

Further enhancement is achieved with $Baseline+ODA+FDA$, primarily attributed to the mitigation of gradient decay. The culmination, $Baseline+MDAT$, showcases performance surges across all datasets, underscoring the synergistic effect of harnessing similarities across domains and the dynamic label smoothing strategy. Collectively, these results attest to the potency of the multi-source domain adversarial training method in extracting domain-neutral features, leading to superior adaptability across diverse domains.

\subsubsection{\textbf{Impact of the Cross-domain Anomaly-aware Contrastive Learning Method}}

To ascertain the contributions of the CACL method, we juxtapose three network configurations:
\begin{itemize}
    \item (d) $Baseline+MDAT+PCL$: Incorporates pixel-level contrastive loss \cite{ICCV_less} based on (c) using random sampling.
    \item (e) $Baseline+MDAT+CPCL$: Employs cross-domain pixel-level contrastive loss.
    \item (f) $Baseline+MDAT+CACL$: Replaces the random sampling strategy with the anomaly-aware sampling strategy, constructing a comprehensive multi-granularity cross-domain alignment framework.
\end{itemize}
The outcomes, presented in Table \ref{tab:ab}, demonstrate substantial performance increments with our method, emphasizing the merits of cross-domain contrastive learning. The $Baseline+MDAT+CPCL$ model, particularly, showcases notable gains in AP on real datasets, while significantly curbing FPR95 across all datasets. This underscores the advantages of aligning sample distributions across domains.

Despite the distinct style differences in the FS Static dataset, the CPCL method registers modest AP gains. Introducing the anomaly-aware sampling strategy, the $Baseline+MDAT+CACL$ model consistently excels in all metrics (barring FPR95 on FS Static). This reaffirms our assertion that the anomaly-aware sampling strategy effectively tempers the influence of redundant easy samples, fostering a more efficient execution of cross-domain pixel-level contrastive loss.

\section{Conclusion}
This work introduced the Multi-Granularity Cross-Domain Alignment (MGCDA) framework, the solution tailored to mitigate the domain discrepancies inherent between synthetic training datasets and real-world testing scenarios. \textit{At the heart of this framework are two pivotal modules: the Multi-source Domain Adversarial Training (MDAT) and the Cross-domain Anomaly-aware Contrastive Learning (CACL).} The MDAT module, with its multi-source domain adversarial loss and dynamic label smoothing mechanism, effectively harnesses multi-source training data. This allows for domain adversarial training across multiple levels, ensuring that the model's features are more robust and domain-agnostic. On the other hand, the CACL method utilizes an anomaly-aware sampling strategy, ensuring efficient pixel-level contrastive learning across domains. This not only aligns the representations of samples from various domains but also significantly bolsters the model's resilience.
\textit{Notably, our MGCDA framework has demonstrated substantial improvements on two real-world datasets, without necessitating additional inference parameters.} This streamlined nature coupled with its efficacy underscores its potential to be seamlessly integrated into a myriad of anomaly segmentation architectures. As the domains of synthetic and real-world data continue to converge, frameworks such as MGCDA will be indispensable in bridging the divide, ensuring models that are both accurate and universally applicable.

\begin{acks}
This research was supported by several funding sources. We thank the National Natural Science Foundation of China (Grant Nos. 62372387, 62001400) for partial financial support. In addition, we acknowledge contributions from the Key R\&D Program of Guangxi Zhuang Autonomous Region, China (Grant Nos. AB22080038, \\AB22080039), the Institute of Applied Physics and Computational Mathematics, Beijing (Grant No. HXO2020-118), the China Postdoctoral Science Foundation (Grant No. 2021M702713), and the Fundamental Research Funds for the Central Universities (Grant Nos. 2682022JX007, 2682022KJ044, 2682022KJ056) were invaluable.

In addition, we acknowledge the support provided to Zhi-Qi Cheng by the U.S. Department of Transportation (69A3551747111). We also acknowledge the significant support that Zhi-Qi Cheng received from the Intel and IBM Fellowships, which were instrumental in advancing his research contributions to this work.
\end{acks}

\bibliographystyle{ACM-Reference-Format}
\balance
\bibliography{AnomalySeg}
\end{document}